%% file: nips_2016.tex
\documentclass{article}

\PassOptionsToPackage{numbers, compress}{natbib}
\usepackage[utf8]{inputenc} 
\usepackage[T1]{fontenc}    
\usepackage{hyperref}       
\usepackage{url}            
\usepackage{booktabs}       
\usepackage{amsfonts}       
\usepackage{nicefrac}       
\usepackage{microtype}      
\usepackage[final]{nips_2016}
\usepackage{hyperref}

\usepackage{lipsum} 
\usepackage{amsmath}
\usepackage{amsfonts}
\usepackage{algorithm}
\usepackage{algorithmicx}
\usepackage{algpseudocode}
\usepackage{graphicx}
\usepackage{subfigure}
\usepackage{multirow}
\usepackage{hhline}

\title{Hierarchical Attention Network for Action Recognition in Videos}

\author{
  Yilin Wang\\
  Arizona State University\\
  \texttt{ywang370$@$asu.edu} 
  \And
  Suhang Wang\\
  Arizona State University\\
  \texttt{suhang.wang$@$asu.edu}\\
   \And
   Jiliang Tang \\
   Yahoo Research \\
   \texttt{jlt$@$yahoo-inc.com} \\
   \AND
   Neil O'Hare \\
   Yahoo Research \\
   \texttt{nohare$@$yahoo-inc.com} \\
   \And
   Yi Chang \\
   Yahoo Research \\
   \texttt{yichang$@$yahoo-inc.com} \\
   \And
    Baoxin Li\\
    Arizona State University\\
    \texttt{baoxin.li$@$asu.edu} 
}

\begin{document}

\maketitle

\begin{abstract}
Understanding human actions in wild videos is an important task with a broad range of applications. In this paper we propose a novel approach named Hierarchical Attention Network (HAN), which enables to incorporate static spatial information, short-term motion information and long-term video temporal structures for complex human action understanding. Compared to recent convolutional neural network based approaches, HAN has following advantages -- (1) HAN can efficiently capture video temporal structures in a longer range;
(2) HAN is able to reveal temporal transitions between frame chunks with different time steps, i.e. it explicitly models the temporal transitions between frames
as well as video segments and (3) with a multiple step spatial temporal attention mechanism, HAN automatically learns important regions in video frames and temporal segments in the video. The proposed model is trained and evaluated on the standard video action benchmarks, i.e., UCF-101 and HMDB-51, and it significantly outperforms the state-of-the arts.
\end{abstract}

\input{introduction}

\input{proposed}

\input{experiment}

\input{conclusion}


\subsubsection*{References}

\bibliographystyle{plain}
\begingroup
\renewcommand{\section}[2]{}%
\bibliography{nips.bib}
\endgroup
\end{document}

%% file: introduction.tex
\section{Introduction}

Understanding human actions in wild videos can advance many real-world applications such as social activity analysis, video surveillance and event detection.   Earlier works typically rely on hand craft features to represent videos~\cite{schuldt2004recognizing,wang2013action}.  They often consist of two steps: motion detection and feature extraction.  First, motion detectors are applied to detect the informative motion regions in the videos and then, hand craft descriptors such HOG ~\cite{dalal2005histograms},  SIFT, or improved Dense Trajectories (iDT)~\cite{wang2013action} extract the feature patterns from those motion regions to represent the video.   In contrast to hand-craft shallow video representation, recent efforts try to learn video representation automatically from large scale labeled video data~\cite{karpathy2014large,simonyan2014two,donahue2015long,wang2015action,ji20133d}.  For example, In~\cite{karpathy2014large}, authors stack the video frames as the input for convolution neural networks (CNN) and two stream CNNs ~\cite{simonyan2014two} combine optical flow and RGB video frames  to train CNN and achieve comparable results with the state-of-the art hand craft based methods.  Very recently, dense trajectory pooled CNN that combines iDT and two stream CNNs via the pooling layer achieves the state-of-the-art performance. However, ~\cite{simonyan2014two} and ~\cite{wang2015action} merely use short term motions that cannot capture the order of motion segments and semantic meanings.

The challenges of action recognition in wild videos are three-fold. First, there are large intra-class appearance and motion variances in the same action due to different viewpoints, motion speeds, backgrounds, etc. Second, wild videos are often collected from movies, TV shows and media platforms and usually have very low resolutions and noise background clutters, which exacerbate the difficulty for video understanding. Third, long range temporal dependencies are very difficult to capture. For example, the Optical Flow,  iDT and 3D ConvNets ~\cite{ji20133d} are computed within a short-time window.  Long Short Term Memory (LSTM) has been recently applied to video analysis \cite{donahue2015long} that provides a memory cell for long temporal information. However, it has been shown that the favorable time range of LSTM is around 40 frames ~\cite{sharma2015action,yue2015beyond}.  In this work, we aim to develop a novel framework to tackle these obstacles for action recognition in videos.  

\begin{figure}[t]
	\centering
	\includegraphics[width=6in]{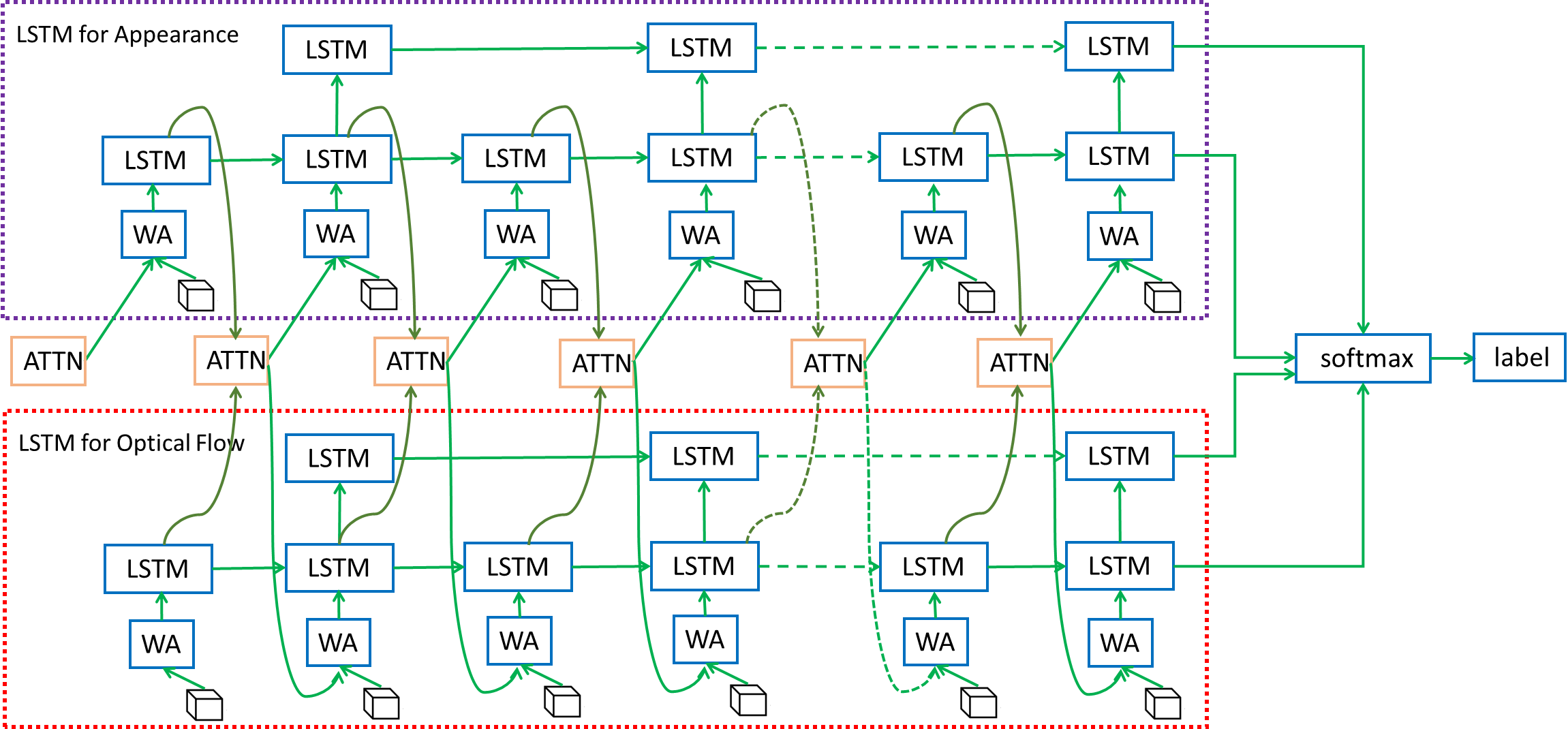}
	\caption{An Illustration of the Proposed Model. The block LSTM means a LSTM cell, whose structure is given in Figure \ref{fig:lstm_one_cell}. The block ATTN indicates the operation to calculate attention weights by using the encoded features from both LSTMs. The block WA represents the weighted average of the input features with the weights from ATTN.}
	\label{fig:hierarchical_two_stream_LSTM_with_attention}
\end{figure}

In this paper, we study the problem of video representation learning for action recognition. In particular, we investigate -- (1) how to utilize the temporal structures in the video to handle intra-appearance variances and background clutters by capturing the informative spatial regions; and (2) how to model the short-term as well as long-term motion dependencies for action recognition.  Providing answers to these two research questions, we propose a novel Hierarchical Attention Network (HAN) that employs a hierarchical structure with recurrent neural network unit e.g., LSTM and a soft spatial temporal attention mechanism for video action recognition.  Our contributions can be summarized as below:
\begin{itemize}
	\item We propose a novel deep learning framework HAN for video action recognition which can explicitly capture both short-term and long-term motion information in an end to end process.
	\item A soft attention is adopted on the  spatial-temporal input features with LSTM to learn the important regions in a frame and the crucial frames in the videos.
	\item  We conduct an extensive set of experiments to demonstrate that the proposed framework HAN is superior to  both state-of-the art  shallow video representation based approaches and deep video representation based approaches on benchmark datasets.
\end{itemize}
The rest of this paper is organized as follows. Section 2 reviews related works. Section 3 describes the proposed hierarchical attention 
deep learning framework in detail. Experimental results and comparisons are discussed in Section 4, followed by conclusions in Section 5.

\section{Related Work}

\textbf{Hand crafted features}:  Video action recognition is a longstanding topic in computer vision community. Many hand-crafted features are used in still images. For example, ~\cite{schuldt2004recognizing} extends 2D harri corner detector to spatial temporal 3D interest points and achieves good performance with SVM classifier. Later, HOG3D that is based on HOG feature \cite{dalal2005histograms} shows its effectiveness by using integral images.  Then improved Dense trajectories ~\cite{wang2013action} has dominated the filed of action recognition. It densely samples interest points and tracks them within a short time period. For each point, local descriptors such as HOG, MBH and HOF are extracted for representation. Then, all the features are encoded by the fish vector as the final video representation. 

\textbf{Deep learned features}:  Deep learning such as convolutional neural network has been shown its success in object detection and image classification in recent years. Based on image CNN, ~\cite{ji20133d,karpathy2014large} extends the CNN framework to videos by stacking video frames. However, the performance is lower than iDT based approaches.  In order to better incorporate temporal information, ~\cite{simonyan2014two} proposes two stream CNNs and achieves comparable results with the state-of-the art performance of hand craft based feature representations \cite{wang2013action}.  To consider the time dependency, \cite{yue2015beyond} proposes a LSTM based recurrent neural network to model the video sequences. Different from \cite{yue2015beyond} that uses a stack based LSTM, the proposed HAN proposes a hierarchical structure to model the video sequences in a multi-scale fashion.  

\textbf{Recurrent visual attention model}:  Visual attention model aims to capture the property of human perception mechanism by identifying the interesting regions in the images. Saliency detection \cite{itti1998model}  typically predicts the human eye movement and fixations for a given image. In \cite{xu2015show}, a recurrent is proposed  to understand where the model focuses on image caption generations. Moreover, recurrent attention model has been applied to other sequence modeling such as machine translation \cite{luong2015effective} and image generation\cite{gregor2015draw}.

%% file: proposed.tex
\section{The Proposed Method}
The overall architecture of HAN is shown in ~\ref{fig:hierarchical_two_stream_LSTM_with_attention}. We describe the three major components of HAN in this section -- the input appearance and motion CNN feature extraction, temporal sequence modeling and hierarchical attention model. 

\subsection{Appearance and Motion Feature Extraction}

In general, HAN can adopt any deep convolution networks~\cite{ji20133d,simonyan2014two,wang2015action} for feature extraction. In this paper, we use two stream ConvNets ~\cite{simonyan2014two} to extract both appearance and motion features.  Specifically, we train a VGG net \cite{simonyan2014very} to extract feature map $f_{P}$ and $f_{Q}$ for the $t$-th frame image $P^{t}$ and the corresponding optical flow image $Q^{t}$:
\begin{equation}
\begin{aligned}
f_{P^{t}}=CNN_{vgg}(P^{t}) \\
f_{Q^{t}}=CNN_{vgg}(Q^{t}) 
\end{aligned}
\end{equation}

Unlike two stream ConvNets \cite{simonyan2014two} that employs the last fully connected layer as the input feature, we use the features for $f_{P}$ and $f_{Q}$ from the last convolutional layer after pooling, which contains the spatial information of the input appearance and motion images.  The input appearance and motion images are first rescaled to $224 \times 224$ and the extracted feature maps from the last pooling layer have the dimension of $D\times K\times K$ ($512\times 14\times14$ used in VGG net).  $K\times K$  is the number of regions in the input image and $D$ is the number of the feature dimensions.  Thus at each time step, we extract $K^2D$ dimension feature vectors for both appearance and motion images.  We refer these feature vectors as feature cube shown in Figure ~\ref{fig:attention}. Then, the feature maps $f_{P^t}$ and $f_Q$ can be denoted in matrix forms as $\mathbf{P}^{t} = [\mathbf{p}_1 ^{t}, \mathbf{p}_2^{t}, \dots, \mathbf{p}_{K^2}^t ] \in \mathbb{R}^{D \times K^2}$  and $\mathbf{Q}^{t} = [\mathbf{q}_1 ^{t}, \mathbf{q}_2^{t}, \dots, \mathbf{q}_{K^2}^t ] \in \mathbb{R}^{D \times K^2}$, respectively.
 
\subsection{Recurrent Neural Network}
Long-short term memory (LSTM), which has the ability to preserve sequence information over time and capture long-term dependencies, has become a very popular model for sequential modeling tasks such as speech recognition~\cite{graves2013hybrid}, machine translation~\cite{bahdanau2014neural} and program execution~\cite{zaremba2014learning}. Recent advances in computer vision also suggest that LSTM has potentials to model videos for action recognition~\cite{sharma2015action}. We follow the LSTM implementation in~\cite{zaremba2014recurrent}, which is given as follows

\begin{figure}[h]
    \centering
    \includegraphics[width=110mm,height=70mm]{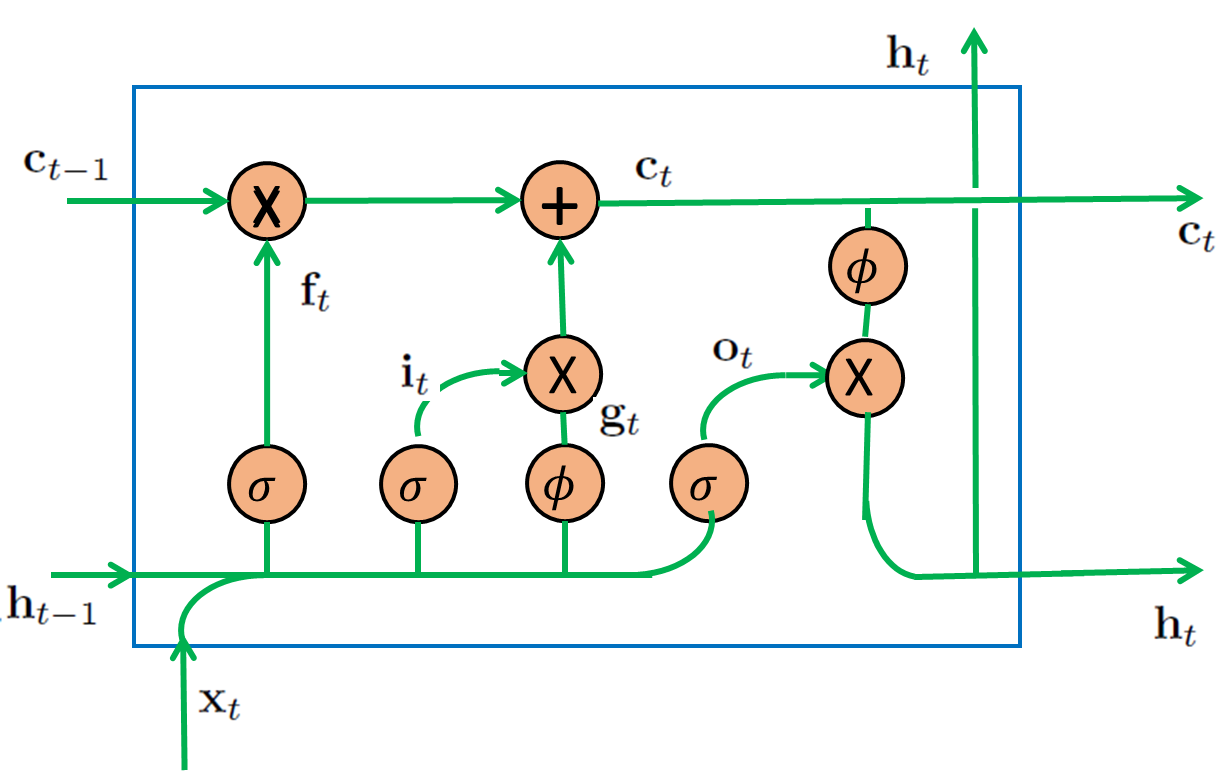}
    \caption{Illustration of One LSTM Cell}
    \label{fig:lstm_one_cell}
\end{figure}

\begin{align}
\mathbf{i}_t & = \sigma(\mathbf{W}_{ix} \mathbf{x}_t + \mathbf{W}_{ih} \mathbf{h}_{t-1} + \mathbf{b}_i) \nonumber \\
\mathbf{f}_t & = \sigma(\mathbf{W}_{fx} \mathbf{x}_t + \mathbf{W}_{fh} \mathbf{h}_{t-1} + \mathbf{b}_f) \nonumber \\
\mathbf{o}_t & = \sigma(\mathbf{W}_{ox} \mathbf{x}_t + \mathbf{W}_{oh} \mathbf{h}_{t-1} + \mathbf{b}+o) \nonumber \\
\mathbf{g}_t & = \tanh(\mathbf{W}_{gx} \mathbf{x}_t + \mathbf{W}_{gh} \mathbf{h}_{t-1} + \mathbf{b}_g) \nonumber \\
\mathbf{c}_t & = \mathbf{f}_t \odot \mathbf{c}_{t-1} + \mathbf{i}_t \odot \mathbf{g}_t \nonumber \\
\mathbf{h}_t & = \mathbf{o}_t \odot \tanh(\mathbf{c}_t)
\end{align}
where $\mathbf{i}_t$ is the input gate, $\mathbf{f}_t$ is the forget gate, $\mathbf{o}_t$ is the forget fate, $\mathbf{c}_t$ is the memory cell state at $t$ and $\mathbf{x}_t$ is the input features at t. $\sigma(\cdot)$ means the sigmoid function and $\odot$ denotes the Hadmard product. The main idea of the LSTM model is the memory cell $\mathbf{c}_t$, which records the history of the inputs observed up to $t$. $\mathbf{c}_t$ is a summation of -- (1) the previous memory cell $\mathbf{c}_{t-1}$ modulated by a sigmoid gate $\mathbf{f}_t$, and (2) $\mathbf{g}_t$, a function of previous hidden states and the current input modulated by another sigmoid gate $\mathbf{i}_t$. The sigmoid gate $\mathbf{f}_t$ is to selectively forget its previous memory while $\mathbf{i}_{t}$ is to selectively accept the current input. $\mathbf{i}_t$ is the gate controlling the output. The illustration of a cell of LSTM at the time step $t$ is shown in Figure \ref{fig:lstm_one_cell}. Next we will introduce how to model both appearance and  motion features using LSTM and how to integrate attention by using encoded features.

\subsection{Hierarchical LSTMs to Capture Temporal Structures}
One natural way of modeling videos is to feed features $\mathbf{P}^t$ and $\mathbf{Q}^t$ into two LSTMs and then put a classifier at the output of LSTMs for classification. However, this straightforward method doesn't fully utilize the structure of actions. In real world, an action is usually composed of a set of sub-actions, which means that video temporal structures are intrinsically layered. For example, a video about long jump consists of three sub-actions -- pushing off the board, flying over the pit and landing. As the three actions take place sequentially, there are strong temporal dependencies among them thus we need to appropriately model the temporal structure among the three actions. In the meantime, the temporal structure within each action is composed of multiple actions.  For example, pushing off the board is composed of running and jump. In other words, the actions we want to recognize are layered and we need to model video temporal structure with multiple granularities. However, directly applying LSTM cannot capture this property. To fully capture the video temporal structure, we develop a hierarchical LSTM. An illustration of the hierarchical LSTM is shown in the purple rectangle in Figure \ref{fig:hierarchical_two_stream_LSTM_with_attention}. This hierarchical LSTM is composed of two layers -- the first layer accepts the appearance feature of each frame as the input and the output of the first layer LSTM is used as the input of the second layer LSTM. To capture the dependencies between different sub-actions such as dependencies between pushing off the board, flying over the pit and landing, we skip every $k$ encoded features from LSTM and use that as the input to the second layer. In addition to capturing the video temporal structure, another advantage of layered LSTM is to increase the learning capability of LSTM. By adding another layer in LSTM, we allow LSTM to learn higher level and more complex features, which is a common practice proven to work well in other deep architectures such as CNN, DNN and DBM. Thus, as shown in Figure \ref{fig:hierarchical_two_stream_LSTM_with_attention}, we use two hierarchical LSTMs to model the appearance and motion features, respectively.

\subsection{Attention Model to Capture Spatial Structures}

\begin{figure}[h]
\centering
\includegraphics[width=120mm,height=100mm]{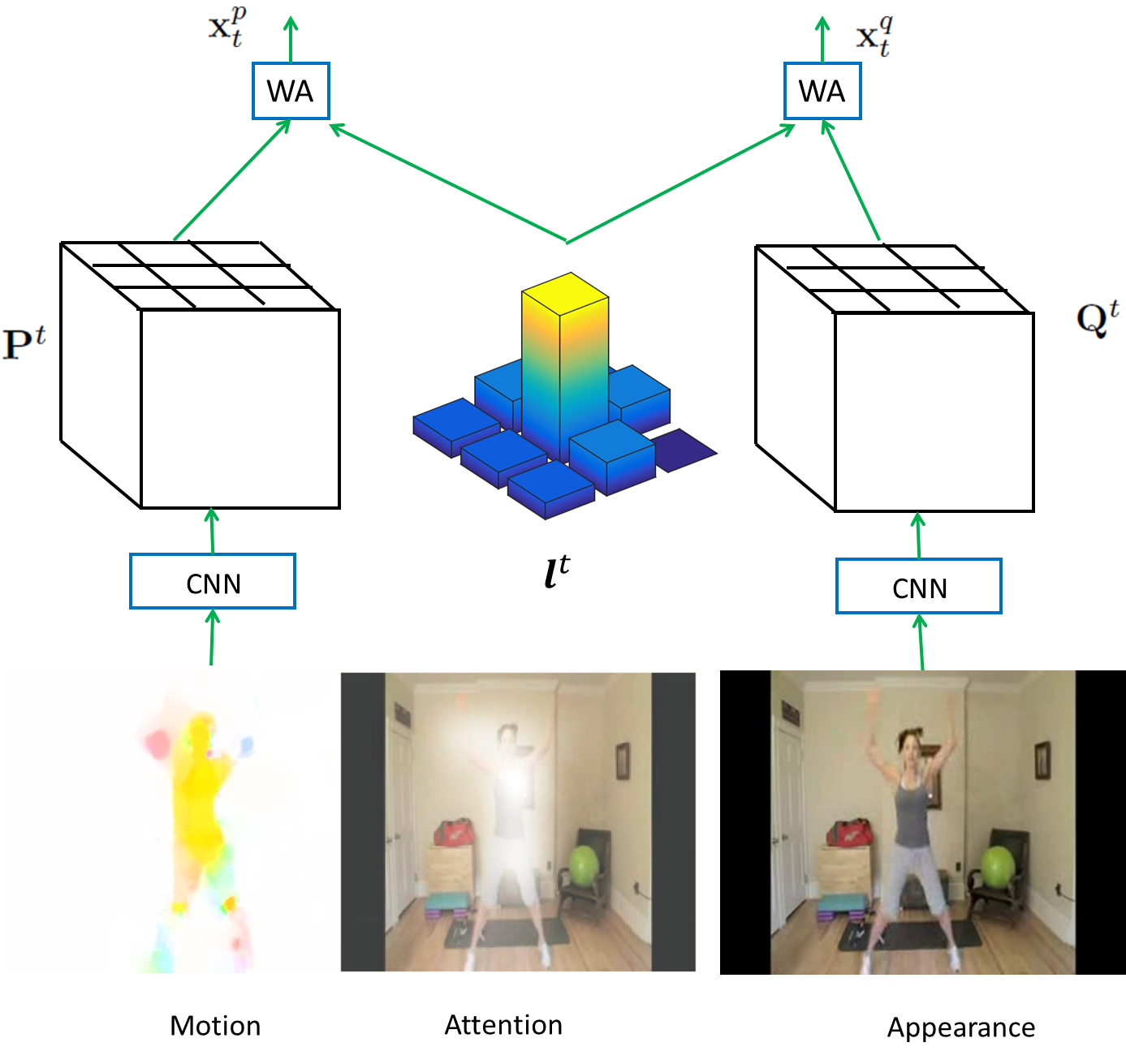}
\caption{attention}
\label{fig:attention}
\end{figure}

The $K^2$ vectors in the appearance $\mathbf{P}^t$ (or motion features $\mathbf{Q}^t$) correspond to $K^2$ regions in the $t$-th frame, which essentially encode the spatial structures. For action recognition, not every region of the frames are relevant for the task at hand. Obviously, we want to focus on the regions where the action is happening. For the action shown in Figure \ref{fig:attention}, we want to mainly focus on hands and legs that are useful for identifying the action; while the background is noisy as a person can perform the same action at different locations. Therefore we could confuse the classifier if we also target on backgrounds. Thus, it is natural for us to assign different attention weights to different regions of the frame. Since video frames are sequential, neighboring frames have strong dependencies, which suggests that we can use the encoded features at time $t-1$ to predict the attention weights at time $t$ and then use the attention weights to refine the input. Specifically, at each time step $t-1$, we use a softmax function over $K \times K$ locations to predict the importance of the $K^2$ locations in the frame, which is written as:
\begin{equation}
	l_i^t = \frac{\exp(\mathbf{w}_i^T \mathbf{h}_{t-1})}{\sum_{j=1}^{K^2} \exp(\mathbf{w}_j^T \mathbf{h}_{t-1})}
\end{equation}
where $l_i^t$ is the importance weight of the $i$-th region of the $t$-th frame, $\mathbf{W} = \{ \mathbf{w}_1, \mathbf{w}_2, \dots, \mathbf{w}_{K^2}\} \in \mathbb{R}^{2D \times K^2}$ are the weights of the softmax function and $\mathbf{h}_{t-1}$ is the concatenation of $\mathbf{h}_{t-1}^{p,1}$ and $\mathbf{h}_{t-1}^{q,1}$, i.e., the encoded appearance and motion features of the $(t-1)$-th frame from the first layer LSTM. Note that we use the encoded appearance feature and motion feature jointly to compute the attention weights instead of computing two attention weights by using the two features, separately. Its advantages are two fold. First, the flow and appearance features capture different aspects of the frame but the attention location on the video should be the same, thus we do not need to calculate two sets of attentions for appearance and optical LSTM separately, which may introduce more computational cost. Second, appearance and motion features provide complimentary information that may help predict more accurate attention. With the attention weights given above, the inputs of the two LSTMs are the weighted average of different locations as:
\begin{equation}
	\mathbf{x}_t^{p} = \sum_{i=1}^{K^2} l_i^t \mathbf{p}_{i}^t ~~~ \text{and} ~~~ \mathbf{x}_t^{q} = \sum_{i=1}^{K^2} l_i^t \mathbf{q}_{i}^t
\end{equation} 

\subsection{Action Recognition with HAN}

We use $\mathbf{h}_T^{p,i}$ to denote the encoding of the the video by the $i$-the layer LSTM for the appearance features and $\mathbf{h}_T^{q,i}$ for motion features. As mentioned above, the hierarchical LSTM captures multi-granularity of video temporal structures, thus, encoded features in different levels (different $i$) provide distinct descriptions of different granularity about actions, which are all useful for action recognition. In addition, the two LSTMs encode complementary information from appearance and motion features, thus encoded features from appearance and motion are also relevant for action recognition. Therefore, we concatenate these features as $\mathbf{h}_f = [\mathbf{h}_T^{p,1}, \dots, \mathbf{h}_T^{p,L}, \mathbf{h}_T^{q,L}, \dots, \mathbf{h}_T^{q,L}]$, where $L$ is the number of layers in each LSTM. We then use the softmax function to predict the probability that the video $v_i$ is classified into the class $c$ as
\begin{equation}
	p(c|v_i, \text{HAN}) = \text{softmax}(\text{HAN}(v_i)) 
\end{equation}
and the loss function is
\begin{equation}
	\max_{\text{HAN}, \mathbf{W}_s} \sum_{i=1}^N \log p( y_i |v_i, \text{HAN})
\end{equation}
where $N$ is number of videos, $y_i$ is the label of $v_i$ and $\mathbf{W}_s$ are the weights of the softmax classifier.


%% file: experiment.tex
\section{Experiments}
In this section, we first present the details of datasets and the evaluation protocol. Then, we describe the details of the implementation of our method. Finally, we present the experimental results with discussions.

\subsection{Datasets and evaluation protocol}

The evaluation is conducted on two public benchmark datasets, i.e.,  UCF-101~\cite{soomro2012ucf101} and HMDB51\cite{kuehne2011hmdb}.  These two datasets are among the largest available annotated video action recognition datasets that have been used in~\cite{karpathy2014large,sharma2015action,simonyan2014two,wang2013action,wang2015action}. Specifically, UCF-101 contains $~13,000$ videos annotated into $101$ action classes with each class having at least 100 videos.  HMDB51 is composed of $~6,700$ videos from $51$ action categories and each category has at least 100 video clips.  For both datasets, the evaluation protocol is the same -- we follow the train/test splits provided by the corresponding organizers. The performance is measured by the mean of accuracies across all the splits in each dataset. 

\subsection{Experiments Setting}

\textbf{Training two stream CNNs and HANs:}\quad Compared to image classification and detection, training a good deep convolutional neural network for videos understanding is more challenging.  Similar to \cite{wang2015action,simonyan2014two}, we use the training data in UCF 101 split to train two stream CNNs. In our implementation, we use the Caffe toolbox \cite{jia2014caffe} and the layer configuration is the same as \cite{simonyan2014very}. All hidden layers use the rectification activation functions and max pooling is performed over $2\times 2$.  Finally, each of the two networks contains 13 convolutional layers and 3 fully connected layers.  The training procedure is similar to \cite{wang2015action,simonyan2014two}, where we use mini-batch stochastic gradient descent with momentum (0.9).  The learning rate is initially set to $10^{-2}$, then changed to $10^{-3}$ after $10,000$ iterations and stopped after $30,000$ iterations and $10,000$ iterations for spatial and temporal nets, respectively. We use the Theano toolbox for HAN implementation and the model is trained by using Adadelta ~\cite{zeiler2012adadelta}. The dimension of LSTM is $1,024$ and the batch size is fixed to $128$ . Techniques of dropout \cite{dahl2013improving} and BPTT  are used. 

\textbf{Optical flow:}\quad The optical flow is computed by the off-the-shelf OpenCV toolbox with GPU implementation of \cite{zach2007duality}. Since the computational cost of optical flow is the bottleneck for the two stream CNN training. We pre-computed all the optical flow images and stored the horizontal and vertical components. The optical flow is computed by the adjacent two frames.  In the testing stage, we fix the number of frames with the equal temporal window between them. 

\subsection{ Results and Analysis}
\begin{figure}[!ht]
\centering
	\subfigure[The sampled frame]{
		\label{fig:att_orig}
		\includegraphics[width=40mm,height=35mm]{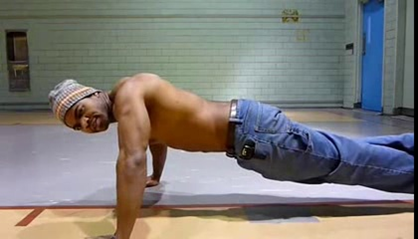}}
	\subfigure[Attention results from HAN]{
		\label{fig:att_pro}
		\includegraphics[width=40mm,height=35mm]{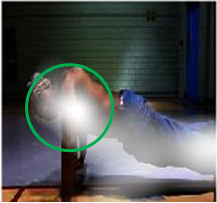}}
	\subfigure[Attention results from \cite{sharma2015action}]{
		\label{fig:att_other}
		\includegraphics[width=40mm,height=35mm]{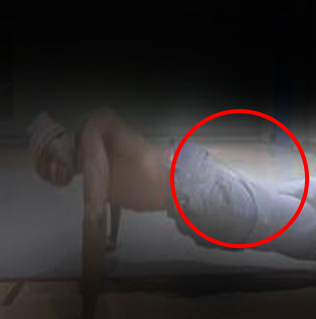}} \\
	\subfigure[The sampled frame]{
		\label{fig:att1_orig}
		\includegraphics[width=40mm,height=35mm]{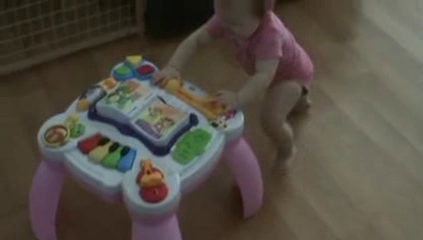}}
	\subfigure[Attention results from HAN]{
		\label{fig:att1_pro}
		\includegraphics[width=40mm,height=35mm]{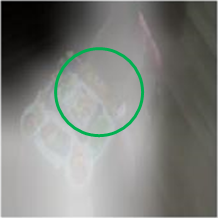}}
	\subfigure[Attention results from \cite{sharma2015action}]{
		\label{fig:att1_other}
		\includegraphics[width=40mm,height=35mm]{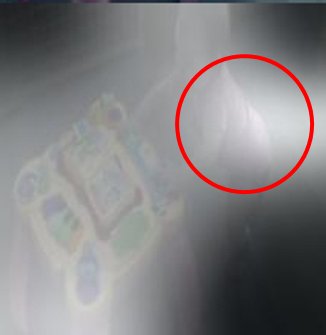}}
	\caption{Visual attention comparison between HAN and soft attention model in \cite{sharma2015action}, the green and red circles highlight the most important region learned by HAN and \cite{sharma2015action} respectively.}
	\label{fig:comparison}
\end{figure}

We compare our models with a set of baselines proposed recently ~\cite{sharma2015action,yue2015beyond,simonyan2014two,wang2015action,karpathy2014large,peng2014action}  including shallow video representation methods and deep ConvNets methods. 
\begin{table}[!h]
	\caption{Average accuracy over three splits on UCF-101 and HMDB51}
	\label{tab: HAN}
	\begin{tabular}{c||c||c} 
		\textbf{Model} & \textbf{UCF-101} & \textbf{HMDB51} \\ \hline\hline
		Full HAN (spatial CNN cube+temporal CNN cube)  &92.7\%	&  64.3\% \\ \hline
		HAN without attention$^{1}$(spatial CNN cube +temporal CNN cube) & 90.6\% & 62.0\%  \\\hline
		HAN without attention$^{2}$(spatial CNN 4096+ temporal CNN 4096) &  91.1\% & 62.7\% \\ \hline\hline
		Spatial HAN (spatial CNN cube) & 75.1\% &  47.7\%\\\hline
		Temporal HAN (temporal CNN cube) & 85.4\% &58.3\% \\\hline\hline
	\end{tabular}
\end{table}
We first evaluate our proposed HAN on UCF-101 and HMDB51 datasets by comparing HAN with different settings to show the importance of each key component in HAN in Table ~\ref{tab: HAN}.  Then, we further compare HAN with state-of-the art methods and experimental results are reported in Table~\ref{tab: state-of-the art}. From the tables, we can make the following observations:

\begin{itemize}
	\item  The proposed method with hierarchical LSTM outperforms methods without hierarchical structures \cite{yue2015beyond,karpathy2014large,simonyan2014two}. These results support that (1) the usage of LSTM can capture video sequences by considering the order of the motion transitions; and (2) the proposed hierarchical structure can effectively model the complex and long time range actions in videos. 
	\item Compared with methods without the attention components, the proposed  HAN encourages the model to focus on the important regions in frames during the learning process, which improves the discriminative ability for classification. For example, in Figure ~\ref{fig:comparison} (b) and Figure ~\ref{fig:comparison} (e), we can see that our model can learn the  important regions for actions more accurately.   
	\item The temporal and spatial features are complementary.  First, by combining them together, both of them have been improved significantly. Second, compared with \cite{sharma2015action} that only considers attention in spatial, HAN can predict more motion related regions in the videos. Third, compared to TDD, the proposed HAN achieves comparable results without considering the iDT information, which suggests that the learned attention regions can have the similar ability to dense trajectory points and reduce the negative impact of background noises. 
	\item Compared to state-of-the art methods on UCF and HMDB51, HAN outperforms them remarkably except \cite{wang2015action}.  The major reason for the exception is that the dataset HMDB is relatively small and the content is unconstrained, while the method in~\cite{wang2015action} incorporates iDT features that are computationally expensive. 
\end{itemize}

\begin{table}[t]
	\centering
	\caption{Comparison with state of the art methods on UCF101 and HMDB51.}
	\label{tab: state-of-the art}
	\begin{tabular}{c || c || c} 
		\textbf{Model} & \textbf{UCF-101} & \textbf{HMDB51} \\ \hline\hline
		Histogram of Oriented Gradient  & 72.4 \% & 40.2\% \\\hline
		Improved dense trajectories (iDT) \cite{wang2013action} &	85.9\%& 57.2\% \\ \hline
		iDT + Stack Fish Vector \cite{peng2014action} & N/A & 66.8\% \\\hline\hline
		spatial-temporal CNN \cite{karpathy2014large}  & 65.4\% &N/A \\\hline
		two stream CNN \cite{simonyan2014two} & 88.0\% & 59.4\% \\\hline
		two stream CNN+LSTM \cite{yue2015beyond} & 88.6\% & N/A  \\\hline
		two stream CNN + iDT \cite{wang2015action} & 91.5\%  & 65.9\% \\\hline
		Soft Attention +LSTM \cite{sharma2015action} & 84.96\%  & 41.3\% \\\hline \hline
		Hierarchical Attention Networks  & 92.7\% & 64.3\% \\\hline\hline
		
	\end{tabular}
\end{table}

%% file: conclusion.tex
\section{Conclusion and Future Work}
In this paper, we propose a hybrid deep framework by incorporating a hierarchical structure and joint attention model to the two stream convnet approach for human action recognition.  The experimental results  suggest that the proposed framework outperforms the two stream convnet approach. Despite using the only optical flow images as input, HAN achieves comparable performance with the state-of-the art method TDD that is much more computationally expensive. These results further support that (1) the hierarchical structure in HAN is important because it can model the frame transitions as well as long video segments and (2) the joint visual attention can help HAN focus on the important video regions and reduce the effect of noisy background. HAN is powerful in sequence modeling thus we would like to explore more applications for HAN in the future such as video event detection since a video event usually contains many sub-events and these sub-events have high dependencies to each. 


%% file: nips_2016.bbl
\begin{thebibliography}{10}

\bibitem{bahdanau2014neural}
Dzmitry Bahdanau, Kyunghyun Cho, and Yoshua Bengio.
\newblock Neural machine translation by jointly learning to align and
  translate.
\newblock In {\em Proceedings of ICLR}, 2015.

\bibitem{dahl2013improving}
George~E Dahl, Tara~N Sainath, and Geoffrey~E Hinton.
\newblock Improving deep neural networks for lvcsr using rectified linear units
  and dropout.
\newblock In {\em ICASSP}, pages 8609--8613. IEEE, 2013.

\bibitem{dalal2005histograms}
Navneet Dalal and Bill Triggs.
\newblock Histograms of oriented gradients for human detection.
\newblock In {\em CVPR}. IEEE, 2005.

\bibitem{donahue2015long}
Jeffrey Donahue, Lisa Anne~Hendricks, Sergio Guadarrama, Marcus Rohrbach,
  Subhashini Venugopalan, Kate Saenko, and Trevor Darrell.
\newblock Long-term recurrent convolutional networks for visual recognition and
  description.
\newblock In {\em CVPR}, pages 2625--2634, 2015.

\bibitem{graves2013hybrid}
Alan Graves, Navdeep Jaitly, and Abdel-rahman Mohamed.
\newblock Hybrid speech recognition with deep bidirectional lstm.
\newblock In {\em ASRU Workshop}, pages 273--278. IEEE, 2013.

\bibitem{gregor2015draw}
Karol Gregor, Ivo Danihelka, Alex Graves, and Daan Wierstra.
\newblock Draw: A recurrent neural network for image generation.
\newblock {\em arXiv preprint arXiv:1502.04623}, 2015.

\bibitem{itti1998model}
Laurent Itti, Christof Koch, and Ernst Niebur.
\newblock A model of saliency-based visual attention for rapid scene analysis.
\newblock {\em TPAMI}, (11):1254--1259, 1998.

\bibitem{ji20133d}
Shuiwang Ji, Wei Xu, Ming Yang, and Kai Yu.
\newblock 3d convolutional neural networks for human action recognition.
\newblock {\em TPAMI}, 35(1):221--231, 2013.

\bibitem{jia2014caffe}
Yangqing Jia, Evan Shelhamer, Jeff Donahue, Sergey Karayev, Jonathan Long, Ross
  Girshick, Sergio Guadarrama, and Trevor Darrell.
\newblock Caffe: Convolutional architecture for fast feature embedding.
\newblock In {\em Multimedia}, pages 675--678. ACM, 2014.

\bibitem{karpathy2014large}
Andrej Karpathy, George Toderici, Sanketh Shetty, Thomas Leung, Rahul
  Sukthankar, and Li~Fei-Fei.
\newblock Large-scale video classification with convolutional neural networks.
\newblock In {\em CVPR}, 2014.

\bibitem{kuehne2011hmdb}
Hildegard Kuehne, Hueihan Jhuang, Est{\'\i}baliz Garrote, Tomaso Poggio, and
  Thomas Serre.
\newblock Hmdb: a large video database for human motion recognition.
\newblock In {\em ICCV}. IEEE, 2011.

\bibitem{luong2015effective}
Minh-Thang Luong, Hieu Pham, and Christopher~D Manning.
\newblock Effective approaches to attention-based neural machine translation.
\newblock {\em arXiv preprint arXiv:1508.04025}, 2015.

\bibitem{peng2014action}
Xiaojiang Peng, Changqing Zou, Yu~Qiao, and Qiang Peng.
\newblock Action recognition with stacked fisher vectors.
\newblock In {\em Computer Vision--ECCV 2014}, pages 581--595. Springer, 2014.

\bibitem{schuldt2004recognizing}
Christian Sch{\"u}ldt, Ivan Laptev, and Barbara Caputo.
\newblock Recognizing human actions: a local svm approach.
\newblock In {\em ICPR}, volume~3, pages 32--36. IEEE, 2004.

\bibitem{sharma2015action}
Shikhar Sharma, Ryan Kiros, and Ruslan Salakhutdinov.
\newblock Action recognition using visual attention.
\newblock In {\em Proceedings Workshop of ICLR}, 2015.

\bibitem{simonyan2014two}
Karen Simonyan and Andrew Zisserman.
\newblock Two-stream convolutional networks for action recognition in videos.
\newblock In {\em NIPS}, pages 568--576, 2014.

\bibitem{simonyan2014very}
Karen Simonyan and Andrew Zisserman.
\newblock Very deep convolutional networks for large-scale image recognition.
\newblock {\em arXiv preprint arXiv:1409.1556}, 2014.

\bibitem{soomro2012ucf101}
Khurram Soomro, Amir~Roshan Zamir, and Mubarak Shah.
\newblock Ucf101: A dataset of 101 human actions classes from videos in the
  wild.
\newblock {\em arXiv preprint arXiv:1212.0402}, 2012.

\bibitem{wang2013action}
Heng Wang and Cordelia Schmid.
\newblock Action recognition with improved trajectories.
\newblock In {\em CVPR}, pages 3551--3558, 2013.

\bibitem{wang2015action}
Limin Wang, Yu~Qiao, and Xiaoou Tang.
\newblock Action recognition with trajectory-pooled deep-convolutional
  descriptors.
\newblock In {\em CVPR}, pages 4305--4314, 2015.

\bibitem{xu2015show}
Kelvin Xu, Jimmy Ba, Ryan Kiros, Aaron Courville, Ruslan Salakhutdinov, Richard
  Zemel, and Yoshua Bengio.
\newblock Show, attend and tell: Neural image caption generation with visual
  attention.
\newblock {\em arXiv preprint arXiv:1502.03044}, 2015.

\bibitem{yue2015beyond}
Joe Yue-Hei~Ng, Matthew Hausknecht, Sudheendra Vijayanarasimhan, Oriol Vinyals,
  Rajat Monga, and George Toderici.
\newblock Beyond short snippets: Deep networks for video classification.
\newblock In {\em CVPR}, pages 4694--4702, 2015.

\bibitem{zach2007duality}
Christopher Zach, Thomas Pock, and Horst Bischof.
\newblock A duality based approach for realtime tv-l 1 optical flow.
\newblock In {\em Pattern Recognition}, pages 214--223. Springer, 2007.

\bibitem{zaremba2014learning}
Wojciech Zaremba and Ilya Sutskever.
\newblock Learning to execute.
\newblock {\em arXiv preprint arXiv:1410.4615}, 2014.

\bibitem{zaremba2014recurrent}
Wojciech Zaremba, Ilya Sutskever, and Oriol Vinyals.
\newblock Recurrent neural network regularization.
\newblock {\em arXiv preprint arXiv:1409.2329}, 2014.

\bibitem{zeiler2012adadelta}
Matthew~D Zeiler.
\newblock Adadelta: an adaptive learning rate method.
\newblock {\em arXiv preprint arXiv:1212.5701}, 2012.

\end{thebibliography}
